\documentclass[11pt]{article}

\usepackage[final]{acl}

\usepackage{times}
\usepackage{latexsym}

\usepackage[T1]{fontenc}

\usepackage[utf8]{inputenc}

\usepackage{microtype}

\usepackage{inconsolata}

\usepackage{graphicx}

\usepackage{algorithm}
\usepackage{algorithmic}
\usepackage{amsmath}
\usepackage{amssymb}
\usepackage{amsfonts}

\usepackage{xcolor}
\usepackage{arydshln}
\usepackage{booktabs}

\usepackage{newfloat}
\usepackage{listings}
\usepackage{multirow}

\title{Model-Agnostic Meta Learning for Class Imbalance Adaptation}

\author{Hanshu Rao\\
  University of Memphis \\
  \texttt{hrao@memphis.edu} \\\And
  Guangzeng Han \\
  University of Memphis \\
  \texttt{ghan@memphis.edu} \\\And
  Xiaolei Huang \\
  University of Memphis \\
  \texttt{xiaolei.huang@memphis.edu} \\
  }

\begin{document}
\maketitle
\begin{abstract}
Class imbalance is a widespread challenge in NLP tasks, significantly hindering robust performance across diverse domains and applications. 
We introduce Hardness-Aware Meta-Resample (HAMR)\footnote{Code is available at \url{https://github.com/trust-nlp/ImbalanceLearning}}, a unified framework that adaptively addresses both class imbalance and data difficulty. 
HAMR employs bi-level optimizations to dynamically estimate instance-level weights that prioritize genuinely challenging samples and minority classes, while a neighborhood-aware resampling mechanism amplifies training focus on hard examples and their semantically similar neighbors. 
We validate HAMR on six imbalanced datasets covering multiple tasks and spanning biomedical, disaster response, and sentiment domains. 
Experimental results show that HAMR achieves substantial improvements for minority classes and consistently outperforms strong baselines. 
Extensive ablation studies demonstrate that our proposed modules synergistically contribute to performance gains and highlight HAMR as a flexible and generalizable approach for class imbalance adaptation.

\end{abstract}

\section{Introduction}



Class imbalance naturally exists in a wide range of tasks, such as text classification~\citep{PADURARIU2019Dealing} and named entity recognition~\citep{nemoto2024majority}, yet it remains a challenge for training robust models.
For instance, in sentiment analysis of product reviews, a large volume of generic positive feedback can easily drown out the few highly informative negative reviews that signal critical product flaws~\citep{henning2023survey}.
This overwhelming prevalence of majority classes biases standard models, causing them to achieve high accuracy by simply predicting the dominant class while failing to learn the features of rare but often more critical minority classes. 
Consequently, this leads to poor generalization and a significant drop in performance on real-world data where identifying these minority classes is paramount.



To address data imbalance, previous studies predominantly pursued two complementary strategies, reweighting and sampling, to counteract natural bias toward majority classes. 
The first involves algorithmic modifications that reweight the loss function to penalize errors on minority classes more heavily, such as inverse frequency weighting~\citep{Deqing2010Inverse}, focal loss~\citep{lin2017focal}, and the Dice loss~\citep{li2020dice}. 
The second strategy focuses on data-level interventions that resample the training distribution, ranging from simple oversampling and sophisticated synthetic data generation~\citep{Kevin2011SMOTE,lu2025multiconir,han-etal-2025-attributes}. 
A growing body of work further blends these ideas by calibrating loss terms with the ``effective number'' of samples or by decoupling representation learning from classifier training~\citep{cui2019classbalanced, kang2020decoupling}

\begin{figure*}[hbt]
\centering
\includegraphics[width=0.95
\textwidth]{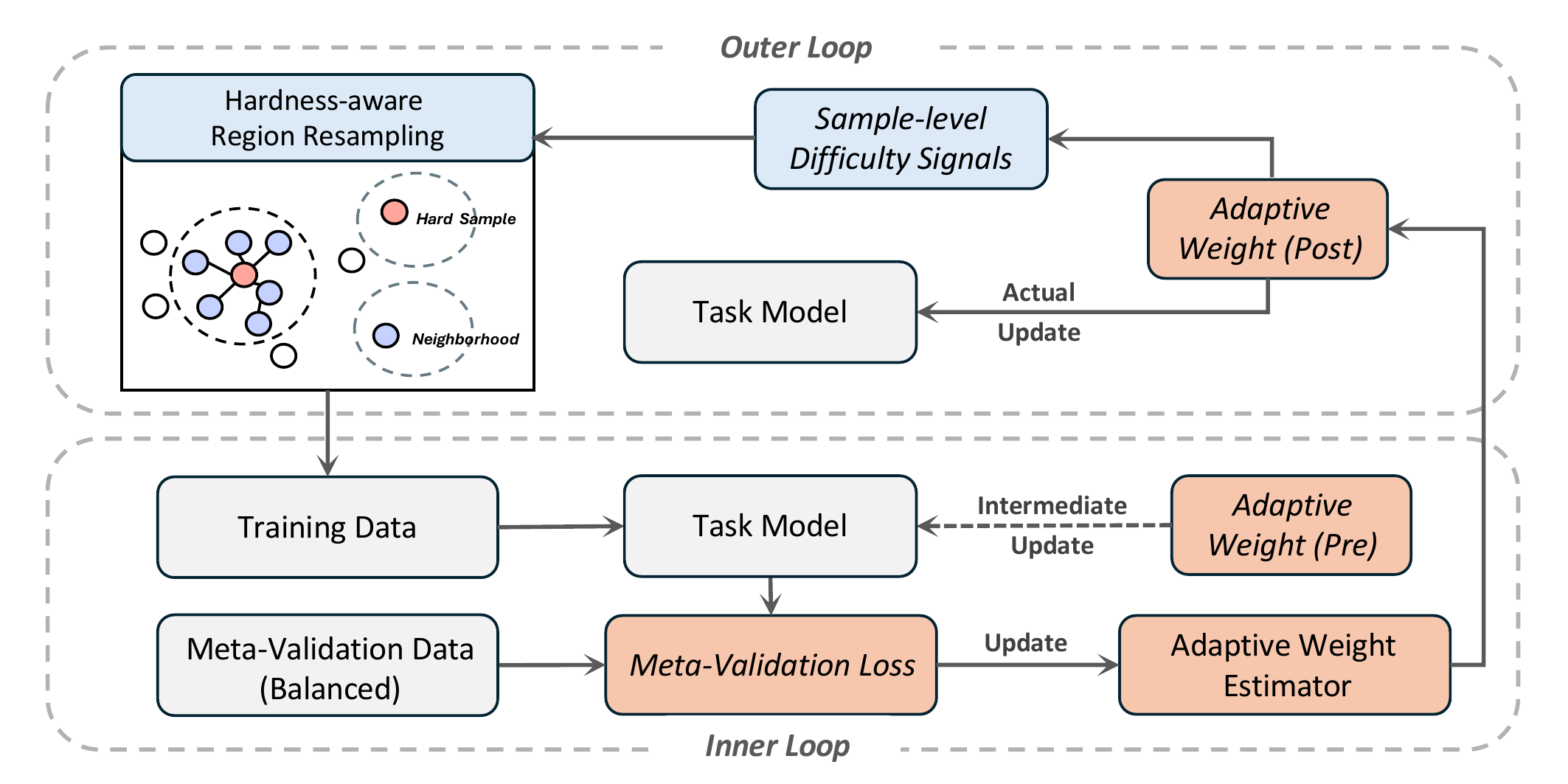} 
\caption{\textbf{Framework of Hardness-Aware Meta-Resample (HAMR).} 
HAMR employs a bi-level optimization: an inner loop performs an intermediate model update using pre-meta weights, and an outer loop updates the weighting network from meta-validation feedback and applies its post-meta weights for the actual model update.
Embedding-based neighborhoods guide resampling toward clusters of hard examples, complementing adaptive weighting to boost generalization.}
\label{Method_Framework_Figure}
\end{figure*}

However, these methods typically rely on pre-defined, static heuristics that treat all samples within a class homogeneously, applying fixed adjustment rates throughout the entire training process~\citep{van2022harm}. 
This uniform treatment overlooks a critical insight: example difficulty does not necessarily align with class membership.
Not all minority class instances are inherently challenging to classify, nor are all majority class examples trivial. 
Consequently, static reweighting schemes may inadvertently down-weight informative majority class examples while over-emphasizing easy minority instances, leading to suboptimal learning dynamics.
A significant research gap thus exists for adaptive methods that can dynamically identify and prioritize genuinely difficult examples regardless of their class affiliation, adjusting their learning strategy in response to the model's evolving understanding of the data distribution~\citep{Vinicius2023Meta, Jain2024Improving}.

To bridge this gap, we propose \textbf{Hardness-Aware Meta-Resample (HAMR)}, a unified meta-learning framework 
that addresses class imbalance by explicitly modeling instance-level difficulty.
Rather than relying on static class-level heuristics, HAMR dynamically prioritizes training entries and adaptively guides the model learning focus that are most critical for generalization.
Specifically, this process integrates \textit{Adaptive Weight Estimation} to adjust the importance of individual training instances
and \textit{Hardness-Aware Region Resampling} to reshape the effective training distribution.
Our contributions are summarized as follows: 
1) We present Hardness-Aware Meta-Resample (HAMR) that jointly addresses class imbalance and instance-level difficulty and dynamically guides learning by identifying and prioritizing samples that are most critical for generalization across both majority and minority classes.
2) We introduce a bi-level meta-optimization that adaptively estimates instance importance based on their performance contribution, which moves beyond frequency-based or fixed reweighting schemes and allows for evolving data difficulty and model learning dynamics.
3) We propose a hardness-aware region resampling paradigm that restructures the effective training distribution at the semantic-neighborhood level rather than at the level of isolated instances, improving robustness under severe long-tailed distributions.



\section{Method}

Existing approaches including loss re-weighting~\citep{li2020dice} or sampling~\citep{Moreo2015Oversampling} commonly focus on minority classes or weigh hard entries higher in a static view or fixed rate.
But emerging yet unresolved questions are: \textit{will minority classes always be prioritized during the learning phase, and if we can dynamically tune our training priorities on data imbalance patterns, will this be better?}
To guide learning objectives and dynamically adjust training focus, we adopt a meta-learning framework~\citep{ren2018learning} and propose our imbalance learning approach in Figure~\ref{Method_Framework_Figure}, \textbf{Hardness-Aware Meta-Resample (HAMR)}.
HAMR decouples \textit{what} the model should focus on from \textit{how} it should learn from two modules:
\textit{Adaptive Weight Estimation} dynamically guides instance-level learning signals by data difficulty across both majority and minority classes;
while \textit{Hardness-Aware Region Resampling} reshapes the training sample distribution toward challenging semantic regions.
Algorithm~\ref{alg:hamr} jointly optimizes adaptive weighting and region resampling towards imbalance generalization in inner and outer learning loops.

\subsection{Adaptive Weight Estimation}
\label{sec:adaptive_weight_estimation}
Static or heuristic sample weights hardly fit the evolving needs of training, in contrast, our module guides what the model should focus dynamically by measuring sample difficulty and learning effectiveness via the weight networks.
To quantify the difficulty and effectiveness, we employ task-specific measures.
For token-level tasks (e.g., NER), we use the sentence-level maximum token loss as $\tilde{\ell}_i = \max_t \ell_{i,t}$, where $\ell_{i,t}$ is the token-level loss in sequence $i$. 
For sequential labeling or classification~\citep{liu-etal-2025-examining} tasks (e.g., reasoning or inference), we deploy the per-example cross-entropy loss. 
Such settings can capture worst-case difficulty while remaining precise for NLP tasks.
Since raw loss values can vary across batches, we apply batch-wise z-score normalization as $\hat{\ell}_i = \frac{\tilde{\ell}_i - \mu}{\sigma}$ to create a consistent scale, where $\mu$ and $\sigma$ are the mini-batch mean and standard deviation.
These normalized scores are processed by a lightweight weight network $f_\theta$ that maps difficulty scores to importance weights $w_i$. 
For numerical stability, the weights are renormalized to a mean of 1 per mini-batch and clipped within a fixed range.

\begin{equation}
  \label{eq:inner}
  \phi' = \phi - \alpha \nabla_{\!\phi} \sum_{i \in \mathcal{B}} w_i^{\text{pre}} \,\tilde{\ell}_i
\end{equation}
\begin{equation}
  \label{eq:outer}
    \theta \leftarrow \theta - \beta \nabla_{\theta}\mathcal{L}_{\mathrm{meta}}(\mathcal{D}_{\mathrm{meta}};\phi')
\end{equation}

We learn the weighting network $\theta$ under a bi-level meta-learning framework that guides how the task model prioritizes training samples, where $\alpha$ and $\beta$ denote the learning rates for the inner and meta updates, respectively.
In the inner loop, for the current mini-batch $\mathcal{B}$, we first compute per-example \emph{pre-meta} weights $\{w_i^{\text{pre}}\}_{i\in\mathcal{B}}$ with the current weight network and perform an intermediate gradient update on the task model $f_{\phi}$, yielding updated parameters $\phi'$ in Eq.~\eqref{eq:inner}.
In the outer loop, we evaluate the temporarily updated model $f_{\phi'}$ on the balanced meta-validation set $\mathcal{D}_{\mathrm{meta}}$, constructed by taking the full validation split and per-class sampling additional examples from the training split until each class reaches the median class count of the validation split.
We then update the weight network parameters $\theta$ by minimizing the meta objective $\mathcal{L}_{\mathrm{meta}}$ in Eq.~\eqref{eq:outer}.
Here, $\mathcal{L}_{\mathrm{meta}}$ denotes the average cross-entropy loss of $f_{\phi'}$ on the balanced meta-validation set $\mathcal{D}_{\mathrm{meta}}$.
After updating $\theta$ in Eq.~\eqref{eq:outer}, we recompute \emph{post-meta} weights $\{w_i^{\text{post}}\}_{i\in\mathcal{B}}$ for the same mini-batch using the updated weight network ($w_i^{\text{post}} = f_{\theta}\!\left(\hat{\ell}_i\right)$), and use them for the actual update of the task model (Eq.~\eqref{eq:main_update}):

\begin{equation}
  \label{eq:main_update}
  \phi \leftarrow \phi - \alpha \nabla_{\!\phi} \sum_{i \in \mathcal{B}} w_i^{\text{post}} \,\tilde{\ell}_i
\end{equation}

\subsection{Hardness-Aware Region Resampling}

Although adaptive weighting addresses how we learn from samples, it does not control which training sample task model can see. 
Existing imbalance approaches~\citep{li2020dice, wu2023gbc, hu2025learning} assume that data with infrequent labels should be valued more throughout the model learning steps, while ignoring the set of hard data entries can shift.
To address this, HAMR introduces a sampling strategy that dynamically adjusts what difficulty training samples the task models should see by leveraging both individual hardness and semantic neighborhood patterns.


Our resampling strategy actively explores neighborhood regions of data space, not just isolated hard examples.
We employ a neighborhood boost mechanism updated periodically (e.g., at the end of each epoch). 
It first identifies hard samples by selecting those with the highest hardness scores $h_i$ (e.g., the top 20\%).
Then, using K-Nearest Neighbors (KNN)~\citep{Zhang2016KNN} on precomputed embeddings, we build $k$ nearest neighbors for each hard sample. 
For efficiency, we construct a FAISS index~\citep{douze2025faiss} for similarity search acceleration. 
The neighborhood boost score $b_i$ for sample $i$ is calculated as its total hit count, i.e., the number of times it appeared in the neighborhoods of hard samples. This score is normalized by its maximum value to lie in $[0, 1]$. 
This mechanism diffuses hardness from individual hard examples to their semantically similar neighbors.

\begin{equation}
  \label{eq:sample_prob}
  p_i \propto (h_i + \varepsilon)^{ {\tau}} \cdot (1 + \lambda\,b_i)
\end{equation}

The final sampling probability $p_i$ integrates each sample's individual hardness $h_i$ with its neighborhood boost $b_i$, applying temperature smoothing ($ {\tau} < 1$) to encourage balanced exploration as shown in Eq.~\eqref{eq:sample_prob}, where $\lambda$ controls the neighborhood boost strength and $\varepsilon$ ensures numerical stability. 
To estimate individual hardness $h_i$, we first use the updated weighting network from \S\ref{sec:adaptive_weight_estimation} to initialize a per-sample weight $w_i$ per training sample.
The global hardness score $h_i$ is updated over time using an exponential moving average (EMA)~\cite{morales2024exponential} of $w_i^{\text{post}}$, where $\gamma$ is the EMA smoothing factor.
EMA smooths noisy parameter updates and acts as a regularization term to reduce noisy effects and improve optimization robustness.
Samples with larger $h_i$ are treated as harder and are assigned higher sampling probabilities when forming training mini-batches.
\begin{equation}
  \label{eq:hardness_update}
  h_i \leftarrow \gamma \cdot h_i + (1-\gamma) \cdot w_i^{\text{post}}
\end{equation}

\begin{algorithm}[t]
\caption{HAMR unified training procedure}
\label{alg:hamr}
\textbf{Input}: dataset $\mathcal{D}$, meta-set $\mathcal{D}_{\text{meta}}$, model $f_{\phi}$, weight network $f_{\theta}$\\
\textbf{Parameters}: learning rates $\alpha,\beta$, EMA factor $\gamma$, KNN refresh interval $F$
\begin{algorithmic}[1]
    \FOR{each epoch}
        \IF{epoch mod $F = 0$}
            \STATE Update neighbourhood boosts $b$ via KNN on hard samples
        \ENDIF
        \STATE Compute sampling probabilities $p$ and sample mini-batch $\mathcal{B}$
        \STATE Compute per-example losses $\tilde{\boldsymbol{\ell}}$ on $\mathcal{B}$; 
        $\mathbf{w}^{\text{pre}} \gets f_{\theta}(\hat{\boldsymbol{\ell}})$
        \STATE \textbf{Inner step:} $\phi' \gets \phi - \alpha \nabla_\phi \langle \mathbf{w}^{\text{pre}}, \tilde{\boldsymbol{\ell}} \rangle$
        \STATE \textbf{Meta step:} $\theta \gets \theta - \beta \nabla_\theta \mathcal{L}_{\text{meta}}(\mathcal{D}_{\text{meta}}; \phi')$; $\mathbf{w}^{\text{post}} \gets f_{\theta}(\hat{\boldsymbol{\ell}})$
        \STATE \textbf{Outer step:} $\phi \gets \phi - \alpha \nabla_\phi \langle \mathbf{w}^{\text{post}}, \tilde{\boldsymbol{\ell}} \rangle$
        \STATE Update $\mathbf{h} \gets \gamma \mathbf{h} + (1-\gamma)\cdot\mathbf{w}^{\text{post}}$
    \ENDFOR
\end{algorithmic}
\end{algorithm}


\subsection{Unified Training Procedure}

The adaptive weighting and region resampling mechanisms integrate seamlessly into a unified training procedure in Algorithm \ref{alg:hamr}.
At each epoch, the framework refreshes neighborhood boosts using KNN, samples data batches according to learned hardness-aware probabilities, and computes adaptive weights reflecting sample hardness.
The core bi-level optimization consists of an inner loop (intermediate model update using pre-meta weights), a meta step (updating the weighting network based on meta-validation feedback), and an outer loop (actual model update using post-meta weights produced by the updated weighting network).

\section{Experiment}

We conduct experiments comparing HAMR with several state-of-the-art baselines on standard NLP benchmarks from multiple domains with varying degrees of class imbalance and compare in Table~\ref{tab:deberta_ner_cls_results}.
Our experiments aim to examine 1) if our approach outperforms baselines on majority and minority classes, 2) if our approach has consistent performance across varying datasets and imbalance settings, and 3) if individual modules of our approach contribute to overall improvements.
In this section, we detail the datasets, baselines, experimental settings, ablation studies, and evaluation metrics. 

\subsection{Data}

We evaluate our approach on six publicly available and imbalanced corpora across two tasks: named-entity recognition (NER) and text classification (CLS).
For NER, BioNLP~\citep{collier-kim-2004-introduction} contains PubMed abstracts annotated with five biomedical entities (e.g., protein and DNA); TweetNER~\citep{ushio-etal-2022-tweet} comprises English tweets labeled with seven entity types including location and product; and MIT‐Restaurant~\citep{liu2013asgard} includes online reviews with eight restaurant-related entities (e.g., price and amenity).
For CLS, Hurricane‐Irma17 and Cyclone‐Idai19~\citep{humaid2020} classify disaster-response tweets into 9 and 10 categories respectively, both showing substantial class imbalance, and SST-5~\citep{socher-etal-2013-recursive} categorizes movie reviews into five ratings.
We quantify class imbalance by the ratio $\mathrm{IR} = \lvert\mathrm{maj}\rvert / \lvert\mathrm{min}\rvert$, ranging from 2.1 (SST-5) to 98.4 (Cyclone-Idai19). 
Table~\ref{tab:dataset_overview} summarizes data statistics with imbalance details in Appendix~\ref{sec:appendix_data}.

\begin{table}[ht]
\centering               
\resizebox{1\columnwidth}{!}{%
\begin{tabular}{l l c l c l l l r}
\multirow{2}{*}{\textbf{Name}} & \multirow{2}{*}{\textbf{Size}} & \multirow{2}{*}{\textbf{\#Classes}} & \multirow{2}{*}{\textbf{Task}} & \multirow{2}{*}{\textbf{Domain}} & \multicolumn{3}{c}{\textbf{Splits}} & \multirow{2}{*}{\textbf{IR}} \\
\cmidrule(lr){6-8}
& & & & & \textbf{Train} & \textbf{Valid} & \textbf{Test} & \\
\midrule
BioNLP & 22,402 & 5 & NER & Biomedical & 16,619 & 1,927 & 3,856 & 33.1 \\
TweetNER & 5,768 & 7 & NER & Social &  4,616 &   576 &   576 &  3.7 \\
MIT-Rest. & 9,181 & 8 & NER & Review  &  6,900 &   760 & 1,521 &  5.1 \\
Irma17 & 9,399 & 9 & CLS & Crisis &  6,579 &   958 & 1,862 & 18.7 \\
Idai19 & 3,933 & 10 & CLS & Crisis &  2,753 &   401 &   779 & 98.4 \\
SST-5 & 11,855 & 5 & CLS & Review &  8,544 & 1,101 & 2,210 &  2.1 \\
\end{tabular}%
}
\caption{Data statistics. IR (Imbalance Ratio) is the ratio of largest class size to smallest class size.}
\label{tab:dataset_overview}
\end{table}

\begin{table*}[htp]
\centering
\setlength{\tabcolsep}{3.5pt}
\resizebox{\textwidth}{!}{%
\begin{tabular}{l cc cc cc cc cc cc}
\multirow{3}{*}{\textbf{Method}} 
& \multicolumn{6}{c}{\textbf{NER}} 
& \multicolumn{6}{c}{\textbf{CLS}} \\
& \multicolumn{2}{c}{\textbf{BioNLP}} 
& \multicolumn{2}{c}{\textbf{MIT‐Restaurant}} 
& \multicolumn{2}{c}{\textbf{TweetNER}} 
& \multicolumn{2}{c}{\textbf{Hurricane‐Irma17}} 
& \multicolumn{2}{c}{\textbf{Cyclone‐Idai19}} 
& \multicolumn{2}{c}{\textbf{SST‐5}} \\
& Macro‐F1 & Micro‐F1 & Macro‐F1 & Micro‐F1 & Macro‐F1 & Micro‐F1 
& Macro‐F1 & Micro‐F1 & Macro‐F1 & Micro‐F1 & Macro‐F1 & Micro‐F1 \\
\midrule
Dice & 70.6 & 74.9 & 80.4 & 81.3 & 59.0 & 62.6 & 70.0 & 73.6 & 58.6 & 77.2 & 56.0 & 56.9 \\
Focal & 69.8 & 74.1 & 79.1 & 80.1 & 58.0 & 61.7 & 70.7 & 73.1 & 63.5 & 81.1 & 55.9 & 56.2 \\
ICF & 68.0 & 70.6 & 79.5 & 80.1	& 57.2 & 61.2 & 72.7 & 74.4	& 63.8 & 80.4 &	56.3 & 56.6 \\
EN	& 69.2 & 74.0 & 78.5 & 79.1	& 53.9 & 56.9 & 70.0 & 73.3	& 61.5 & 78.3 & 55.9 & 56.3 \\
GBC & 68.4 & 72.4 & 78.7 & 79.3 & 57.0 & 60.2 & 68.8 & 71.9 & 60.0 & 79.9 & 52.9 & 53.9 \\
LNR & 70.1 & 74.6 & 79.4 & 80.6 & 59.0 & 62.7 & 71.0 & 73.4 & 63.4 & 78.8 & 55.7 & 56.4 \\
HAMR (Ours)  & \textbf{72.7} & \textbf{75.4} & \textbf{81.1} & \textbf{82.1} & \textbf{60.2} & \textbf{63.7} 
               & \textbf{73.4} & \textbf{75.2} & \textbf{65.7} & \textbf{81.4} & \textbf{57.0} & \textbf{57.7} \\
\end{tabular}%
}
\caption{Overall performance (\%) by macro- and micro-F1 scores across 6 datasets.  {Numbers are averaged over three runs with different random seeds.} Best scores are in \textbf{bold}.}
\label{tab:deberta_ner_cls_results}
\end{table*}




\subsection{Baselines}

To validate our approach, we compare HAMR with six state-of-the-art imbalance-learning baselines, each tackling class imbalance through a different mechanism.
\textit{Dice Loss}~\citep{li2020dice} and \textit{Focal Loss}~\citep{lin2017focal} reformulate the training objective by optimizing the Sørensen-Dice coefficient with overlap-based normalization or introducing a modulating factor $(1-p_t)^\gamma$ to dynamically down-weight well-classified instances, respectively.
\textit{Inverse Class Frequency (ICF)}~\citep{he2009Learning}, \textit{Effective Number (EN) Weights}~\citep{cui2019classbalanced}, and \textit{Gradient-based Clustering (GBC)}~\citep{wu2023gbc} follow the re-weighting direction. 
ICF scales the cross-entropy loss of each class by the inverse of its empirical frequency to amplify minority class contributions.
EN reduces returns of additional samples through class-balanced weights $w_y = \tfrac{1-\beta}{1-\beta^{n_y}}$ where $n_y$ is the class frequency. 
GBC is a close meta-learning method that performs weighted $k$-means clustering in gradient space to supervise adaptive loss re-weighting.
Our approach differs from GBC; 
instead of using meta re-weighting alone, we further introduce region-aware resampling to prioritize hard regions during training.
Label-Noise Rebalancing (LNR)~\citep{hu2025learning} is a close study in data augmentation by deliberately flipping a subset of majority-class labels into minority classes without discarding data or synthesizing samples.
We provide more detailed baseline configurations in Appendix~\ref{sec:appendix_baseline}.

\subsection{Experimental Settings}

To ensure consistency with baselines, we use DeBERTa-v3-base~\citep{he2021debertav3} as the default task model while examining decoder-only model options such as Qwen3~\citep{yang2025qwen3} (Sec.~\ref{subsec:base}). 
We report Micro-/Macro-F1 under the standard protocols; for NER, scores are computed at the span level under the BIO scheme excluding the O tag. 
All reported results are averaged over three independent runs with different random seeds.
The standard deviations of the results in Table~\ref{tab:deberta_ner_cls_results} are provided in Appendix~\ref{sec:appendix_std}.
Additional implementation details, hyperparameter sensitivity analyses, and computational overhead are provided in Appendix~\ref{sec:appendix_implementation}, Appendix~\ref{sec:hyperparameter_sensitivity}, and Appendix~\ref{sec:computational}, respectively.

\section{Results}
This section  discusses overall results and ablation studies. 
We conduct four ablation studies to answer the following questions: 
1) does each module of our HAMR approach contribute effectively to better performance? 
2) has our approach successfully encountered imbalance effects on minority classes? 
3) what role does neighborhood expansion play in HAMR?
4) will our approach be generalizable on other base models, such as ModernBERT~\citep{modernbert} and Llama 3~\citep{grattafiori2024llama}?

\begin{table*}[ht]
\centering
\renewcommand{\arraystretch}{1.15}
\setlength{\tabcolsep}{3.5pt}
\resizebox{\textwidth}{!}{%
\begin{tabular}{l cc cc cc cc cc cc}
\multirow{3}{*}{\textbf{Method}} 
& \multicolumn{6}{c}{\textbf{NER}} 
& \multicolumn{6}{c}{\textbf{CLS}} \\
& \multicolumn{2}{c}{\textbf{BioNLP}} 
& \multicolumn{2}{c}{\textbf{MIT‐Restaurant}} 
& \multicolumn{2}{c}{\textbf{TweetNER}} 
& \multicolumn{2}{c}{\textbf{Hurricane‐Irma17}} 
& \multicolumn{2}{c}{\textbf{Cyclone‐Idai19}} 
& \multicolumn{2}{c}{\textbf{SST‐5}} \\
& Macro‐F1 & Micro‐F1 & Macro‐F1 & Micro‐F1 & Macro‐F1 & Micro‐F1 
& Macro‐F1 & Micro‐F1 & Macro‐F1 & Micro‐F1 & Macro‐F1 & Micro‐F1 \\
\midrule
Full & 72.7 & 75.4 & 81.1 & 82.1 & 60.2 & 63.7 & 73.4 & 75.2 & 65.7 & 81.4 & 57.0 & 57.7 \\
w/o AWE & 71.2 & 74.9 & 80.8 & 81.8 & 57.1 & 60.9 &71.3 & 73.7 & 64.1 & 80.7 & 56.5 & 57.5 \\
w/o Resampling & 72.3 & 75.3 & 80.6 & 81.6 & 59.6 & 63.5 & 71.8 & 74.2 & 61.5 & 79.9 & 56.5 & 57.2 \\
w/o KNN & 71.7 & 75.1 & 80.9 & 81.8 & 59.2 & 63.4 & 71.0 & 73.4 & 64.6 & 80.7 & 56.5 & 57.2 \\
w/o All & 71.0 & 74.3 & 79.7 & 80.8 & 57.4 & 60.2 & 70.0 & 72.7 & 61.3 & 79.6 & 55.6 & 56.6 \\
\end{tabular}%
}
\caption{Ablation of individual modules in our HAMR. \textbf{w/o AWE } disables adaptive weight estimation, \textbf{w/o Resampling} removes hardness-aware region resampling, \textbf{w/o KNN} removes KNN-based neighborhood boost, and \textbf{w/o All} removes all modules.}
\label{tab:module_ablation}
\end{table*}

\begin{figure*}[t]
\centering
\includegraphics[width=0.986\textwidth]{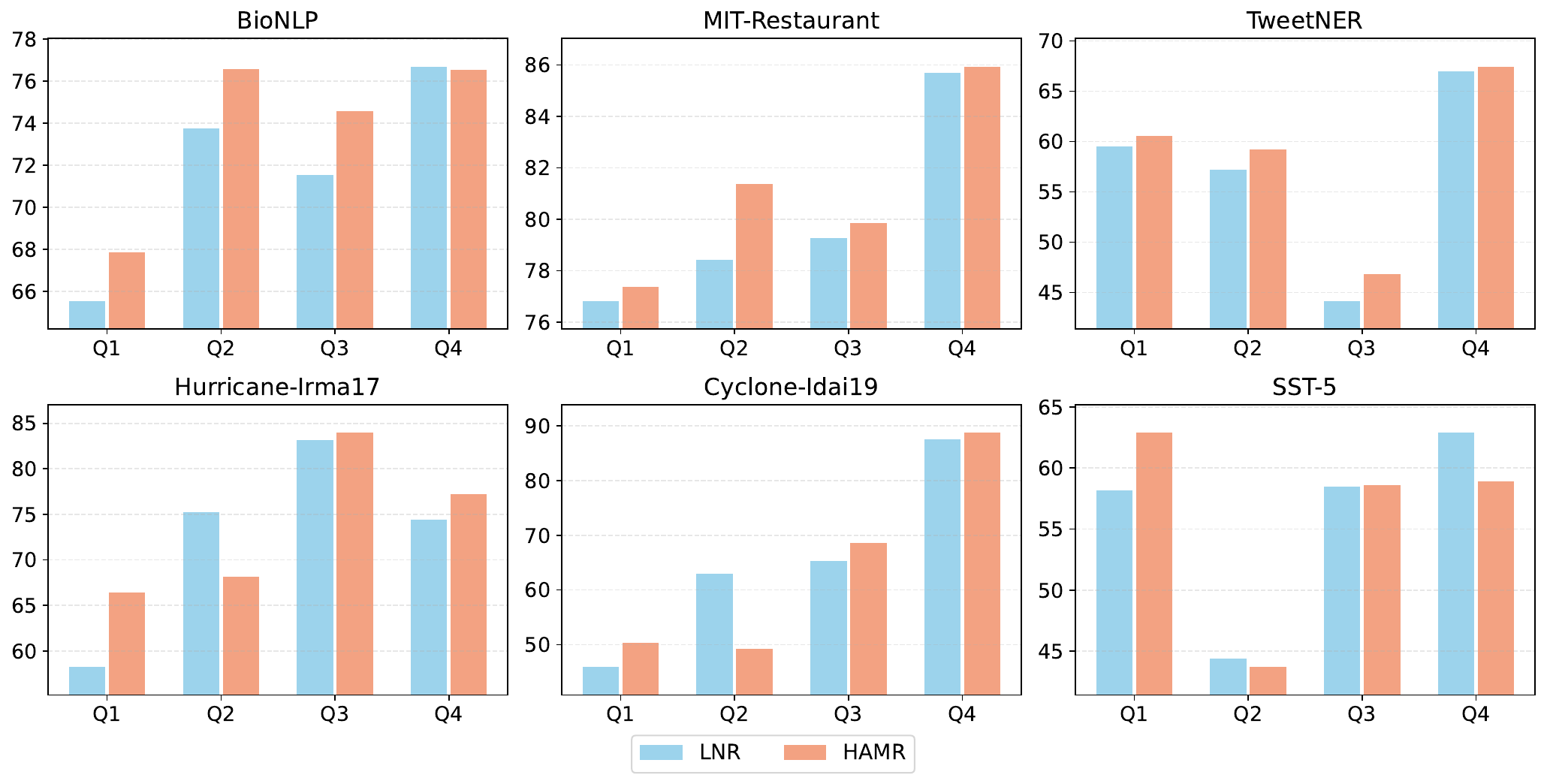} 
\caption{Model performance on majority and minority classes, grouped by quartiles, with Q1 denoting the rarest labels. 
Bars report the mean of per-label F1 scores within each quartile.
}
\label{Minority_Class}
\end{figure*}

\subsection{Overall performance}

Table~\ref{tab:deberta_ner_cls_results} reports the performance of HAMR and six baseline methods on six benchmark datasets with varying imbalance ratios.
Overall, HAMR consistently outperforms all baselines across datasets, achieving absolute gains of 0.7–7.1 percentage points in Macro-F1 and 0.3–6.8 percentage points in Micro-F1.
The improvements are particularly notable on datasets with severe class imbalance, such as Cyclone-Idai19 with an imbalance ratio of 98.4.
Importantly, HAMR also maintains competitive performance on less skewed datasets.
For example, on TweetNER, HAMR achieves a Macro-F1 score of 60.2, compared to 59.0 for Dice and LNR, and 57.0 for the meta-learning-based GBC method.
These results indicate that HAMR can automatically adapt its imbalance learning behavior across both highly imbalanced and relatively balanced settings.

HAMR also shows consistent gains over baselines that address imbalance with different mechanisms. 
On Hurricane-Irma17, HAMR improves over the re-weighting baseline ICF (Macro-F1 73.4 vs. 72.7; Micro-F1 75.2 vs. 74.4).
On Cyclone-Idai19, HAMR surpasses the data-level rebalancing method LNR (Macro-F1 65.7 vs. 63.4; Micro-F1 81.4 vs. 78.8).
On TweetNER, HAMR exceeds the meta-reweighting approach GBC (Macro-F1 60.2 vs. 57.0; Micro-F1 63.7 vs. 60.2).
These comparisons suggest that combining adaptive weighting with region-aware resampling yields more consistent improvements than using a single strategy.
Moreover, the consistent gains observed across both sequence-level and entity-level tasks confirm the generality of our approach.
To further analyze the contribution of individual components, we next present a module-level ablation study.

\subsection{Ablation Study 1: Modules}

We conduct a component-level ablation study by selectively disabling key modules in HAMR:
Adaptive Weight Estimation (``w/o AWE''), Hardness-Aware Region Resampling (``w/o Resampling''), the KNN-based region propagation mechanism (``w/o KNN''), and a full ablation that removes all modules (``w/o All'').
Table~\ref{tab:module_ablation} summarizes the resulting F1 scores across all datasets.

Across all benchmarks, the complete HAMR model consistently yields the strongest performance, indicating that the proposed components are mutually complementary.
Removing all modules (``w/o All'') typically yields the weakest performance, and the full HAMR model brings consistent gains over this vanilla baseline, e.g., improving Macro-F1 from 61.3 to 65.7 on Cyclone-Idai19 and from 70.0 to 73.4 on Hurricane-Irma17.
Among single-module ablations, removing Adaptive Weight Estimation causes substantial degradation, particularly under severe class imbalance,
with Macro-F1 decreasing from 72.7 to 71.2 on BioNLP and from 60.2 to 57.1 on TweetNER.
Performance also deteriorates noticeably when Hardness-Aware Region Resampling is removed,
most prominently on disaster-related datasets with extreme imbalance, such as Cyclone-Idai19
(Macro-F1 decreases from 65.7 to 61.5) and Hurricane-Irma17 (from 73.4 to 71.8).
By contrast, disabling the KNN-based region propagation results in smaller yet consistent losses,
for example, reductions of 0.2 and 1.0 Macro-F1 points on MIT-Restaurant and BioNLP, respectively,
suggesting that this component provides complementary refinements rather than primary gains.

\subsection{Ablation Study 2: Minority Class}

A persistent challenge in imbalance learning is to improve performance in rare classes without compromising frequent-class performance.
To examine whether HAMR effectively benefits minority labels, we compare it against LNR, a representative imbalance-aware baseline.
Figure~\ref{Minority_Class} reports the unweighted mean of per-label F1 within label-frequency quartiles, where Q1 contains the rarest 25\% of labels and Q4 the most frequent.
An effective imbalance learning approach is generally expected to improve performance in Q1 and Q2 while remaining competitive results in Q3 and Q4.

As shown in Figure~\ref{Minority_Class}, HAMR consistently yields substantial gains in the lower-frequency quartiles compared to LNR,
without sacrificing performance on frequent classes.
For example, on SST-5, HAMR improves Q1 mean per-label F1 by 4.7 points over LNR.
Similar improvements in Q1 are observed across other datasets, including BioNLP and Cyclone-Idai19,
indicating that HAMR effectively targets the most challenging rare classes.
Meanwhile, performance on Q3 and Q4 remains stable or shows slight improvements,
demonstrating that gains on minority classes do not come at the expense of frequent-class performance.
Overall, these results confirm that HAMR achieves a favorable balance between rare-class improvement and frequent-class stability.

\subsection{Ablation Study 3: Role of Neighborhood Expansion}

In HAMR, neighborhood expansion refers to the resampling step that extends focus from isolated hard samples to their semantic neighbors through the neighborhood boost mechanism.
We hypothesize that neighborhood expansion improves resampling by moving beyond a small set of isolated ambiguous hard points toward a broader, more label-coherent local region around them.
If so, the expanded resampling pool should exhibit higher local label consistency than hard samples alone.

To examine this, we measure local label consistency using the same KNN definition as in HAMR, where neighbors are built on the full training set with $K=10$.
Following Bahri et al.~\cite{Deep2020Bahri}, for each sample we compute the fraction of its $K$ nearest neighbors that share the same label and mark the sample as locally consistent if this fraction exceeds $0.5$.
We then average this indicator over different sampled sets while keeping the neighbor graph fixed.
Specifically, we consider four settings: Random, where $K$ comparison samples are drawn uniformly from the remaining training set; Full Set, where consistency is aggregated over all training samples; Hard Samples, where aggregation is restricted to the hard set identified at the end of epoch 3; and Hard $\cup$ Neighbors, where the hard set is expanded by taking the union of hard samples and their KNN neighbors.

Table~\ref{tab:local_consistency} reports the average local label consistency under the four settings on the three classification datasets.
As shown, Hard $\cup$ Neighbors consistently achieves higher local label consistency than Hard Samples across all three datasets.
For example, on Cyclone-Idai19, consistency increases from 0.504 to 0.567; on Hurricane-Irma17, from 0.505 to 0.546; and on SST-5, from 0.239 to 0.270.
This pattern indicates that the additional samples introduced by neighborhood expansion tend to come from locally more label-consistent areas than hard samples alone.
Moreover, the random baseline remains substantially lower than all KNN-based settings, indicating that the learned embedding space exhibits meaningful local label clustering rather than arbitrary class-frequency effects.

\begin{table}[ht]
\centering
\resizebox{\columnwidth}{!}{
\begin{tabular}{lccc}
\textbf{Setting} & \textbf{Cyclone-Idai19} & \textbf{Hurricane-Irma17} & \textbf{SST-5} \\
\hline
Random & 0.274 & 0.159 & 0.216 \\
Full Set & 0.551 & 0.511 & 0.260 \\
Hard Samples & 0.504 & 0.505 & 0.239 \\
Hard $\cup$ Neighbors & 0.567 & 0.546 & 0.270 \\
\end{tabular}
}
\caption{Local label consistency under different sampling settings.}
\label{tab:local_consistency}
\end{table}

\begin{table*}[ht]
\centering
\renewcommand{\arraystretch}{1.05}
\setlength{\tabcolsep}{3.5pt}
\resizebox{\textwidth}{!}{%
\begin{tabular}{l cc cc cc cc cc cc}
\multirow{3}{*}{\textbf{Method}} 
& \multicolumn{6}{c}{\textbf{NER}} 
& \multicolumn{6}{c}{\textbf{CLS}} \\
& \multicolumn{2}{c}{\textbf{BioNLP}} 
& \multicolumn{2}{c}{\textbf{MIT‐Restaurant}} 
& \multicolumn{2}{c}{\textbf{TweetNER}} 
& \multicolumn{2}{c}{\textbf{Hurricane‐Irma17}} 
& \multicolumn{2}{c}{\textbf{Cyclone‐Idai19}} 
& \multicolumn{2}{c}{\textbf{SST‐5}} \\
& Macro‐F1 & Micro‐F1 & Macro‐F1 & Micro‐F1 & Macro‐F1 & Micro‐F1 
& Macro‐F1 & Micro‐F1 & Macro‐F1 & Micro‐F1 & Macro‐F1 & Micro‐F1 \\
\midrule
Dice & 64.2 & 67.1 & 71.5 & 72.0 & 47.5 & 52.1 & \textbf{70.1} & 73.0 & 42.7 & 52.2 & 49.9 & 54.1 \\
Focal & 62.2 & 66.0 & 71.2 & 72.0 & 44.7 & 49.4 & 68.2 & 72.0 & 65.4 & 80.4 & 52.7 & 55.1 \\
ICF	& 66.3 & 68.8 & 74.5 & 74.8 & 47.4 & 51.7 & \textbf{70.1} & 72.5 & 60.7 & 79.0	& 52.9 & 54.2 \\
EN & 66.6 & 69.2 & 74.9 & 75.6 & 47.7 & 52.0 & 69.0  & 72.2 & 62.5 & 78.8 & 53.1 & 54.8 \\
GBC & 66.9 & 69.8 & 73.9 & 74.5	& 49.6 & 54.1 & 66.7 & 71.1 & 62.8 & 79.1 & 50.6 & 52.9 \\
LNR & 63.4 & 67.8 & 70.9 & 71.7 & 45.9 & 51.0 & 69.2 & 72.6 & 65.7 & \textbf{81.0} & 53.2 & \textbf{55.2} \\
HAMR (Ours)  & \textbf{68.8} & \textbf{71.3} & \textbf{76.1} & \textbf{76.8} & \textbf{52.2} & \textbf{56.9} & \textbf{70.1} & \textbf{73.1} & \textbf{66.8} & \textbf{81.0} & \textbf{54.1} & 54.3 \\
\end{tabular}%
}
\caption{Performance by macro- and micro-F1 (\%) using ModernBERT-base on six datasets. 
} 
\label{tab:modernbert_ablation}
\end{table*}

\begin{table*}[ht]
\centering
\renewcommand{\arraystretch}{1.1}
\setlength{\tabcolsep}{3.5pt}
\resizebox{\textwidth}{!}{%
\begin{tabular}{l l cc cc cc cc cc cc}
\multirow{3}{*}{\textbf{Model}} 
& \multirow{3}{*}{\textbf{Method}} 
& \multicolumn{6}{c}{\textbf{NER}} 
& \multicolumn{6}{c}{\textbf{CLS}} \\
& 
& \multicolumn{2}{c}{\textbf{BioNLP}} 
& \multicolumn{2}{c}{\textbf{MIT‐Restaurant}} 
& \multicolumn{2}{c}{\textbf{TweetNER}} 
& \multicolumn{2}{c}{\textbf{Hurricane-Irma17}} 
& \multicolumn{2}{c}{\textbf{Cyclone‐Idai19}} 
& \multicolumn{2}{c}{\textbf{SST‐5}} \\
& 
& Macro‐F1 & Micro‐F1 & Macro‐F1 & Micro‐F1 & Macro‐F1 & Micro‐F1 & Macro‐F1 & Micro‐F1 & Macro‐F1 & Micro‐F1 & Macro‐F1 & Micro‐F1 \\
\hline
\multirow{2}{*}{Qwen3‐4B} 
&  {LNR} & 
     {36.4} &  {49.6} &  {60.7} &  {61.8} &  {39.0} &  {42.6} &  {70.4} &  {73.0} &  {56.5} &  {78.3} &  {56.4} &  {57.3} \\
& HAMR & 
    \textbf{40.1}  &  \textbf{51.4}  &  \textbf{64.4} & \textbf{65.5}  & \textbf{40.1} & \textbf{43.5} & \textbf{71.0} & \textbf{74.1} & \textbf{58.3} & \textbf{82.0} & \textbf{56.5} & \textbf{58.2} \\
\hline
\multirow{2}{*}{Qwen3‐8B} 
&  {LNR}   & 
     {37.9} &  {50.4} &  {62.0} &  {63.6} &  {39.1} &  {43.2} &  {69.7} &  {72.9} &  {52.0} &  {78.7} &  {56.3} &  {57.2} \\
& HAMR &  
    \textbf{40.0} & \textbf{52.6} & \textbf{65.2} & \textbf{66.6} & \textbf{40.0} & \textbf{44.3} & \textbf{73.7} & \textbf{74.9} & \textbf{58.8} & \textbf{80.2} & \textbf{56.4} & \textbf{58.2} \\
\hline
\multirow{2}{*}{LLaMa3‐8B} 
&  {LNR}  & 
     {38.0} &  {50.9} &  {63.1} &  {64.3} &  {43.0} &  {47.1} &  {69.6} &  {72.6} &  {58.3} &  {79.1} &  {56.0} &  {57.7} \\
& HAMR &  
    \textbf{42.5} & \textbf{54.7} & \textbf{66.5}  &  \textbf{67.9}  & \textbf{43.4} & \textbf{47.7} & \textbf{71.2} & \textbf{73.7} & \textbf{59.3} & \textbf{80.9} & \textbf{56.8} & \textbf{58.9}\\
\end{tabular}%
}
\caption{Performance by macro- and micro-F1 (\%) using three decoder-only models on NER and CLS datasets.}
\label{tab:decoder_ablation}
\end{table*}

\subsection{Ablation Study 4: Base Model}
\label{subsec:base}

We investigate whether HAMR generalizes across different backbone architectures, including both encoder-based and decoder-only language models.
To this end, we evaluate HAMR on newly released  ModernBERT-base~\citep{modernbert} model,
as well as decoder-only models including Qwen3-4B, Qwen3-8B~\citep{yang2025qwen3}, and LLaMA3-8B~\citep{grattafiori2024llama}.

\paragraph{Encoder-based models.}
Table~\ref{tab:modernbert_ablation} shows that replacing DeBERTa with ModernBERT-base leads to performance reductions across all methods,
indicating that absolute task performance remains sensitive to the representational capacity of the underlying encoder.
Despite this shift, HAMR continues to outperform or match competing approaches across datasets,
achieving the best results on 5 out of 6 benchmarks and yielding notable improvements over strong baselines such as GBC and LNR.
For example, HAMR improves Macro-F1 by 2.6 points over GBC on TweetNER and by 5.4 points over LNR on BioNLP.
These results suggest that the effectiveness of HAMR is robust to changes in encoder architectures.

\paragraph{Decoder-only models.}
For decoder-only architectures, we fine-tune Qwen3-4B, Qwen3-8B, and LLaMA3-8B using LoRA~\citep{Edward2021LoRA},
and compare HAMR against LNR, the strongest baseline in our main experiments.
As reported in Table~\ref{tab:decoder_ablation}, HAMR consistently outperforms LNR across all evaluated settings.
The improvements are most pronounced on highly imbalanced datasets; for instance, on Cyclone-Idai19 (IR = 98.4),
HAMR increases Macro-F1 by 1.8 points with Qwen3-4B and by 6.8 points with Qwen3-8B.
Overall, HAMR yields larger gains on named entity recognition tasks than on text classification tasks,
with median improvements of 3.2 Macro-F1 points for NER and 1.0 point for CLS,
indicating that the method is particularly effective for structured prediction under severe token-level imbalance.

\paragraph{Encoder vs. decoder observations.}
We observe a clear architectural distinction between encoder-based and decoder-only models.
While decoder-only models achieve performance comparable to encoders on classification tasks,
they exhibit weaker results on named entity recognition.
For example, although LLaMA3-8B with HAMR attains competitive performance on Hurricane-Irma17,
its Macro-F1 on BioNLP is lower by more than 30 points,
reflecting the limitations of unidirectional context for sequence labeling.
Overall, these findings confirm that HAMR generalizes well across model architectures and scales,
while delivering the strongest benefits when paired with encoder-based backbones for structured prediction tasks.

\section{Related Work}


\paragraph{Meta Learning} is a framework of ``learning to learn'' and optimizes a learner's update rule, initialization, or data-usage policy via a higher-level meta-objective, so that a few inner updates yield better generalization~\citep{finn2017model}.
Common approaches to generalize models are to learn how to weight or sample training data entries~\citep{ren2018learning, heck2024learning, wu2023gbc}.
\citeauthor{ren2018learning} adjust per-example (or per-batch) weights so that inner-loop steps improve a held-out objective;
\citeauthor{heck2024learning} develop a small network or rule maps training signals (loss/margin) to weights data samples and control the weighting process by a meta-loss;
and \citeauthor{Jain2024Improving} optimize selection of the validation set used in learned reweighting training to improve classifier generalization on images.

However, few studies leverage meta learning on the data imbalance, a critical while widely existing issue in NLP tasks~\citep{liu2020survey, henning2023survey,zhang2025features}.
A close study developed a meta reweighting approach via gradient space clustering that constructs representative meta-sets and clusters per-sample gradients to supervise loss reweighting~\citep{wu2023gbc}.
But our approach differs from the existing work~\citep{Zhang2016KNN, ren2018learning, Jain2024Improving} with a particular focus on the data imbalance~\citep{wu2023gbc} by unifying `how to weight' and `what to sample' under the meta-optimized look, which complements the prior class-level reweighting and pointwise meta-sampling.

\paragraph{Imbalance Learning} for NLP tasks learns to handle data imbalance issues by correcting majority bias, improving minority performance, and calibrating under unequal priors~\citep{johnson2019survey, liu2020survey, henning2023survey}, which has three primary directions, loss re-engineering or class re-weighting~\citep{li2020dice, cao2019ldam, lin2017focal} and data augmentation~\citep{edunov2018understanding, chen2023empirical, song2024toward, hu2025learning}.
For example, \citeauthor{song2024toward} introduces two samplers for imbalanced labels and generates minority-augmented instances with high diversity to augment infrequent classes;
\citeauthor{zhao2023imbalanced} aligns the distributions of feature representations and label representations to alleviate the impact of the distribution gap between the training set and the test set caused by the imbalance issue;
and LDAM \citep{cao2019ldam} dynamically adjusts the learning process to emphasize harder classes.

Our study differs from and complements existing imbalance learning approaches~\cite{lin2017focal, cao2019ldam, li2020dice, zhao2023imbalanced, wu2023gbc, hu2025learning} by introducing an end-to-end meta-learning framework with neighborhood-based resampling.
Our meta-learning approach accumulates hardness over training and propagates it to semantic neighbors to drive resampling and sampling-based rebalancing.
Recent studies show that not all minority samples are equally informative, and some majority examples remain hard to learn and underused~\citep{cao2019ldam, van2022harm, Jain2024Improving, song2024toward}.
We meta-learn example importance to prefer ``hard-but-useful'' samples (even in head classes), and resample semantic neighborhoods so that training focus flows to difficult manifolds.

\section{Conclusion}


We presented Hardness-Aware Meta-Resample (HAMR), a framework that jointly addresses class imbalance and instance-level difficulty in NLP through adaptive sample weighting and semantic neighborhood-based resampling.
HAMR leverages bi-level meta-learning to dynamically adjust sample importance and incorporates semantic neighborhood information to focus training on challenging regions of the data space.
Experiments on six NER and text classification benchmarks demonstrate that HAMR consistently improves performance over strong baselines, with more pronounced gains under severe class imbalance.
Quartile analysis further shows that HAMR enhances performance on infrequent classes without sacrificing performance on frequent ones.
Ablation studies confirm that both adaptive weighting and neighborhood-based resampling contribute to these improvements.
Overall, HAMR provides a flexible and effective approach for mitigating class imbalance across diverse tasks and model architectures.

\section*{Limitations}

Despite its effectiveness, this study has several limitations that suggest directions for future research. 
The proposed HAMR framework involves a bi-level optimization process and periodic neighborhood updates, which introduce additional computational cost compared to conventional single-loop training. 
Although the overhead remains manageable in our experimental settings, scalability to very large models or corpora may require further optimization. 
Moreover, the neighborhood-based resampling depends on the quality of precomputed embeddings, and its performance may degrade when semantic representations are unstable or domain coverage is limited. Another limitation lies in the experimental scope. 
The current evaluation focuses on text classification and named entity recognition tasks, and thus the generalizability of HAMR to other NLP problems, such as relation extraction or multi-modal learning, remains to be verified. 
These aspects merit further investigation toward improving the efficiency, robustness, and applicability of HAMR across broader settings.

\section*{Acknowledgment}
The authors thank anonymous reviewers for their insightful feedback. 
The project was partially supported by the National Science Foundation (NSF) under awards IIS-2245920.
We thank the computing resources provided by the iTiger GPU cluster~\cite{sharif2025ITIGER} supported by the NSF MRI program under the award CNS-2318210.

\bibliography{custom}

\appendix

\section{Data Imbalance Details}
\label{sec:appendix_data}



\paragraph{BioNLP} contains five entity types—DNA, RNA, Protein, Cell-type and Cell-line (Table~\ref{tab:bionlp}) from PubMed abstracts \citep{collier-kim-2004-introduction}. Entities appear in only about 3\% of tokens, and the class-frequency ratio between the most and least common entity types is 33.1, marking an extreme imbalance that stresses biomedical NER systems.
    
\paragraph{MIT-Restaurant} has 7,136 restaurant-search queries with eight domain-specific slots (e.g., Cuisine, Price, Hours) (Table~\ref{tab:mit_restaurant}) \citep{liu2013asgard}. Despite its modest size, skew across the eight classes (IR = 5.1) and short utterance length test a model's ability to exploit limited context.

\paragraph{TweetNER} comprises time-stamped English tweets labeled with seven entity types (Table~\ref{tab:tweetner}) \citep{ushio-etal-2022-tweet}. Social-media noise, rapid topic drift and an IR of 3.7 make it a realistic low-resource setting for NER in the wild.

\begin{table}[htp]
\centering
\small
\begin{tabular}{lrrrr}
\textbf{Label} & \textbf{Train} & \textbf{Val} & \textbf{Test} & \textbf{Overall} \\
\hline
\noalign{\vskip 1.5pt}
Protein & 27240 & 3029 & 5067 & 35336 \\
DNA & 8273 & 1260 & 1056 & 10589 \\
Cell\_type & 6090 & 628 & 1921 & 8639 \\
Cell\_line & 3325 & 505 & 500 & 4330 \\
RNA & 820 & 131 & 118 & 1069 \\
\end{tabular}
\caption{BioNLP Dataset Distribution}
\label{tab:bionlp}
\end{table}

\begin{table}[htp]
\centering
\small
\begin{tabular}{lrrrr}
\textbf{Label} & \textbf{Train} & \textbf{Val} & \textbf{Test} & \textbf{Overall} \\
\hline
\noalign{\vskip 1.5pt}
Location & 3355 & 462 & 812 & 4629 \\
Cuisine & 2532 & 307 & 532 & 3371 \\
Amenity & 2249 & 292 & 533 & 3074 \\
Name & 1755 & 146 & 402 & 2303 \\
Dish & 1353 & 122 & 288 & 1763 \\
Hours & 871 & 119 & 212 & 1202 \\
Rating & 987 & 83 & 201 & 1271 \\
Price & 655 & 75 & 171 & 901 \\
\end{tabular}
\caption{MIT-Restaurant Dataset Distribution}
\label{tab:mit_restaurant}
\end{table}

\begin{table}[ht]
\centering
\small
\begin{tabular}{lrrrr}
\textbf{Label} & \textbf{Train} & \textbf{Val} & \textbf{Test} & \textbf{Overall} \\
\hline
\noalign{\vskip 1.5pt}
Person & 4666 & 598 & 596 & 5860 \\
Event & 2242 & 256 & 265 & 2763 \\
Product & 1850 & 241 & 220 & 2311 \\
Group & 2242 & 227 & 311 & 2780 \\
Creative\_work & 1661 & 208 & 179 & 2048 \\
Corporation & 1700 & 203 & 191 & 2094 \\
Location & 1259 & 181 & 165 & 1605 \\
\end{tabular}
\caption{TweetNER Dataset Distribution}
\label{tab:tweetner}
\end{table}

\paragraph{Hurricane-Irma17} contains 9,399 annotated tweets of Hurricane Irma with nine humanitarian categories (e.g., Infrastructure and utility damage, Rescue volunteering or donation effort) (Table~\ref{tab:irma17}) \citep{humaid2020}. The moderate class imbalance (IR = 18.7) and social media context test a model's robustness to noisy, informal text with uneven class distributions.

\paragraph{Cyclone-Idai19} has 3,900 annotated tweets from Cyclone Idai (2019) with ten humanitarian categories (e.g., Rescue volunteering or donation effort, Sympathy and support) 
(Table~\ref{tab:idai19}) \citep{humaid2020}. Despite its smaller size, the severe class imbalance (IR = 98.4) tests a model's robustness to extreme distributional skew in crisis communication scenarios.

\paragraph{SST-5} is fine-grained Stanford Sentiment Treebank with 11,855 movie-review sentences and a five-point scale from \textit{very negative} to \textit{very positive} (Table~\ref{tab:sst5}) \citep{socher-etal-2013-recursive}. The minority \textit{very positive} class appears in only ~2\,\% of sentences (IR = 2.1), making SST-5 an established but still imbalanced benchmark for sentiment analysis.

\begin{table}[ht]
\centering
\resizebox{\columnwidth}{!}{%
\begin{tabular}{lrrrr}
\textbf{Label} & \textbf{Train} & \textbf{Val} & \textbf{Test} & \textbf{Overall} \\
\hline
Other relevant information & 1651 & 240 & 467 & 2358 \\
Infrastructure and utility damage & 1317 & 192 & 372 & 1881 \\
Rescue volunteering or donation effort & 1113 & 162 & 315 & 1590 \\
Injured or dead people & 626 & 91 & 177 & 894 \\
Displaced people and evacuations & 528 & 77 & 150 & 755 \\
Not humanitarian & 430 & 63 & 122 & 615 \\
Caution and advice & 429 & 62 & 122 & 613 \\
Sympathy and support & 397 & 58 & 112 & 567 \\
Requests or urgent needs & 88 & 13 & 25 & 126 \\
\end{tabular}%
}
\caption{Hurricane-Irma17 Dataset Distribution}
\label{tab:irma17}
\end{table}

\begin{table}[ht]
\centering
\resizebox{\columnwidth}{!}{
\begin{tabular}{lrrrr}
\textbf{Label} & \textbf{Train} & \textbf{Val} & \textbf{Test} & \textbf{Overall} \\
\hline
Rescue volunteering or donation effort & 1308 & 191 & 370 & 1869 \\
Sympathy and support & 338 & 49 & 95 & 482 \\
Injured or dead people & 303 & 44 & 86 & 433 \\
Other relevant information & 285 & 41 & 81 & 407 \\
Infrastructure and utility damage & 248 & 36 & 70 & 354 \\
Requests or urgent needs & 100 & 15 & 28 & 143 \\
Caution and advice & 62 & 9 & 18 & 89 \\
Not humanitarian & 56 & 8 & 16 & 80 \\
Displaced people and evacuations & 40 & 6 & 11 & 57 \\
Missing or found people & 13 & 2 & 4 & 19 \\
\end{tabular}
}
\caption{Cyclone-Idai19 Dataset Distribution}
\label{tab:idai19}
\end{table}

\begin{table}[ht]
\centering
\small
\begin{tabular}{lrrrr}
\textbf{Label} & \textbf{Train} & \textbf{Val} & \textbf{Test} & \textbf{Overall} \\
\hline
\noalign{\vskip 1.5pt}
Negative & 2218 & 289 & 633 & 3140 \\
Positive & 2322 & 279 & 510 & 3111 \\
Neutral & 1624 & 229 & 389 & 2242 \\
Very positive & 1288 & 165 & 399 & 1852 \\
Very negative & 1092 & 139 & 279 & 1510 \\
\end{tabular}
\caption{SST-5 Dataset Distribution}
\label{tab:sst5}
\end{table}

\section{Baseline Details}
\label{sec:appendix_baseline}



\paragraph{Dice Loss}~\citep{li2020dice} optimizes the Sørensen-Dice coefficient and has demonstrated effective performance on several data imbalance benchmarks~\citep{Carole2017Generalised}. 
Its overlap-based normalization naturally de-emphasizes the majority background, making it robust to skewed class distributions.
We adopt its multi-class extension, defined as $\mathcal{L}_{\text{Dice}} = 1 - \frac{2 \sum_{i} y_{i} \hat{y}_{i} + \varepsilon}{\sum_{i} y_{i} + \sum_{i} \hat{y}_{i} + \varepsilon}$, where $y_i \in \{0,1\}$ is a ground-truth indicator for a given class, and $\hat{y}_i \in [0,1]$ is predicted probability.
We initialize the smoothing constant $\varepsilon$ as $10^{-5}$ and optimize hyperparameters by different tasks and datasets.


\paragraph{Focal Loss}~\citep{lin2017focal} adds a modulating factor to focus on data samples with hard minority classes and down-weight well-classified (easy) instances, which formulates as $\mathcal{L}_{\text{Focal}} = -\alpha(1 - p_t)^{\gamma}\log p_t$, where $p_t$ refers to a predicted probability for the true class.
The approach has achieved leading performance in classification tasks~\citep{aljohani2023novel}.
We follow those practice to set the focusing parameter $\gamma=2$ and adopt a class-balanced weighting schema for $\alpha$.


\paragraph{Inverse Class Frequency (ICF)}~\citep{he2009Learning} is a classic re-weighting strategy that scales the loss of each class by the inverse of its empirical frequency in the training set.
This weighting alleviates the dominance of majority classes by amplifying the contribution of minority classes during optimization.
Formally, for class $c$ with empirical frequency $f_c$, the weight is defined as $w_c = \frac{1/f_c}{\frac{1}{C}\sum_{j=1}^{C} 1/f_j}$, normalized to have mean $1$.
We apply ICF on the cross-entropy loss, i.e., $\mathcal{L}_{\text{ICF}} = -\sum{i} w_{y_i} \log p_{y_i}$, where $y_i$ is the ground-truth label and $p_{y_i}$ is the predicted probability.

\paragraph{Effective Number (EN) Weights}~\citep{cui2019classbalanced} reweight cross-entropy loss using the effective number of samples, which models diminishing returns when more data is added to the same class. 
This weighting naturally increases the contribution of minority classes while suppressing the dominance of frequent ones, making it robust to long-tailed distributions. 
We adopt the class-balanced cross-entropy formulation, defined as $\mathcal{L}_{\text{CB-CE}} = - \sum_{i} w_{y_i} \log p_\theta(y_i \mid x_i)$, where $w_y = \tfrac{1-\beta}{1-\beta^{n_y}}$ with class frequency $n_y$ and hyperparameter $\beta \in [0,1)$. 
Following prior work, we set $\beta=0.9999$ across all tasks.

\paragraph{Gradient-based Clustering (GBC)}~\citep{wu2023gbc} is a meta-reweighting method that replaces heuristic meta-set construction with gradient-space clustering. 
Each sample is represented by concatenating per-sample gradients from a fixed number of randomly sampled layers, and weighted $k$-means is performed in this gradient space to select representative meta samples, which then supervise loss re-weighting.
We set the number of clusters to $M=\lceil \rho N \rceil$ (with $\rho=0.1$ in our experiments), extract gradient features from three layers, use $|\cos|$ distance for clustering, and choose the nearest point to each centroid as the meta set.

\paragraph{Label-Noise Rebalancing (LNR)}~\citep{hu2025learning} has achieved top performance in classification tasks and introduced asymmetric label noise as a plug-and-play data-level remedy for class imbalance.
By deliberately flipping a small, carefully chosen subset of majority-class labels into minority classes, LNR shifts the decision boundary toward rare classes without discarding data or synthesizing new samples. 
We employ LNR source codes and adopt the best-practice settings of noise rate ($\rho=0.15$) and temperature schedule.

\section{Implementation Details}
\label{sec:appendix_implementation}

We implement all methods by HuggingFace Transformers~\citep{wolf_2020_transformers} and PyTorch~\citep{Paszke_2019_PyTorch} and conduct experiments on the same data splits. 
All experiments were conducted on a server with an NVIDIA H100 GPU. 
We fine-tuned the microsoft/deberta-v3-base\footnote{\url{https://huggingface.co/microsoft/deberta-v3-base}} model as the backbone for all NER and classification tasks. 
Our proposed HAMR framework was implemented as a custom subclass of the Hugging Face Trainer, integrating a meta-learning module (WNet) for dynamic sample reweighting and a k-NN mechanism for neighbor boosting. 
The k-NN component used pre-computed sentence embeddings generated by the bge-large-en-v1.5 model\footnote{\url{https://huggingface.co/BAAI/bge-large-en-v1.5}} and was accelerated by a faiss-gpu index\footnote{\url{https://faiss.ai/index.html}} for efficient similarity search.
We used the AdamW optimizer with FP16 mixed-precision for 8 epochs. The task model learning rate, along with the specific hyperparameters for our HAMR framework, were tuned for each dataset to achieve optimal performance. 
For each dataset, we performed multiple runs within predefined hyperparameter search spaces (see Table~\ref{tab:hp_search}) and selected the configuration that achieved the best validation score. The resulting settings used for evaluating the test split are summarized in Table~\ref{tab:hyperparameters_settings}.

\begin{table}[ht]
  \centering
  \small
  \begin{tabular}{ll}
    \textbf{Parameter} & \textbf{Search space} \\
    \hline
    \noalign{\vskip 1.5pt}
    \texttt{hardness\_alpha}   & $[0,1]$ \\
    \texttt{knn\_lambda}       & $[0,1]$ \\
    \texttt{knn\_ratio}        & $[0,1]$ \\
    \texttt{knn\_k}            & $\{5,10,15,20\}$ \\
    \texttt{learning\_rate}    & $\{1,2,3,4,5\}\times10^{-5}$ \\
    \texttt{wnet\_lr}          & $\{1,2,3,4,5\}\times10^{-4}$ \\
    \texttt{meta\_update\_lr}  & $\{1,2,3,4,5\}\times10^{-4}$ \\
  \end{tabular}
    \caption{Hyperparameter search spaces.}
  \label{tab:hp_search}
\end{table}

\begin{table*}[ht]
  \centering
  \resizebox{0.9\textwidth}{!}{
  \begin{tabular}{l*{7}{c}}
    \textbf{Dataset} & \textbf{hardness\_alpha} & \textbf{knn\_lambda} & \textbf{knn\_k} & \textbf{knn\_ratio} & \textbf{learning\_rate} & \textbf{wnet\_lr} & \textbf{meta\_update\_lr} \\
    \hline
    BioNLP           & 0.7 & 0.7 & 15 & 0.5 & $2\times10^{-5}$ & $3\times10^{-4}$ & $2\times10^{-4}$ \\
    MIT-Restaurant   & 0.7 & 0.1 & 10 & 0.1 & $2\times10^{-5}$ & $3\times10^{-4}$ & $2\times10^{-4}$ \\
    TweetNER         & 0.7 & 0.7 & 10 & 0.5 & $2\times10^{-5}$ & $3\times10^{-4}$ & $2\times10^{-4}$ \\
    Hurricane-Irma17 & 0.7 & 0.1 & 10 & 0.1 & $4\times10^{-5}$ & $1\times10^{-4}$ & $2\times10^{-4}$ \\
    Cyclone-Idai19   & 0.7 & 0.1 & 10 & 0.1 & $4\times10^{-5}$ & $1\times10^{-4}$ & $2\times10^{-4}$ \\
    SST-5            & 0.6 & 0.3 & 10 & 0.2 & $2\times10^{-5}$ & $5\times10^{-4}$ & $2\times10^{-4}$ \\
  \end{tabular}
  }
  \caption{Hyperparameter settings for different datasets.}
  \label{tab:hyperparameters_settings}
\end{table*}

\begin{table*}[htbp]
\centering
\resizebox{0.91\textwidth}{!}{
\begin{tabular}{lcccccccccccc}
 & \multicolumn{2}{c}{\textbf{BioNLP}} & \multicolumn{2}{c}{\textbf{MIT-Restaurant}} & \multicolumn{2}{c}{\textbf{TweetNER}} & \multicolumn{2}{c}{\textbf{Hurricane-Irma17}} & \multicolumn{2}{c}{\textbf{Cyclone-Idai19}} & \multicolumn{2}{c}{\textbf{SST-5}} 
\\
\cmidrule(lr){2-3} \cmidrule(lr){4-5} \cmidrule(lr){6-7} \cmidrule(lr){8-9} \cmidrule(lr){10-11} \cmidrule(lr){12-13}
\textbf{Method} & \textbf{Macro} & \textbf{Micro} & \textbf{Macro} & \textbf{Micro} & \textbf{Macro} & \textbf{Micro} & \textbf{Macro} & \textbf{Micro} & \textbf{Macro} & \textbf{Micro} & \textbf{Macro} & \textbf{Micro} \\
\midrule
    Dice  & 0.58 & 0.49 & 0.45 & 0.38 & 0.92 & 0.81 & 0.88 & 0.81 & 1.15 & 0.79 & 0.68 & 0.59 \\
    Focal & 0.62 & 0.55 & 0.51 & 0.44 & 0.95 & 0.84 & 0.91 & 0.83 & 1.32 & 0.91 & 0.75 & 0.62 \\
    ICF   & 0.74 & 0.68 & 0.55 & 0.49 & 1.05 & 0.92 & 1.12 & 0.98 & 1.45 & 1.05 & 0.84 & 0.71 \\
    EN   & 0.82 & 0.75 & 0.68 & 0.58 & 1.18 & 1.05 & 1.25 & 1.10 & 1.68 & 1.18 & 0.92 & 0.82 \\
    GBC   & 0.95 & 0.88 & 0.74 & 0.65 & 1.35 & 1.18 & 1.38 & 1.22 & 1.85 & 1.34 & 1.05 & 0.91 \\
    LNR   & 0.68 & 0.58 & 0.52 & 0.46 & 0.98 & 0.82 & 1.02 & 0.92 & 1.28 & 0.85 & 0.78 & 0.68 \\
    HAMR (Ours) & 0.72 & 0.61 & 0.54 & 0.48 & 0.85 & 0.72 & 1.05 & 0.91 & 1.42 & 1.02 & 0.82 & 0.74 \\
\end{tabular}%
}
\caption{Standard deviation ($\pm \sigma$) for Macro-F1 and Micro-F1 scores over 3 runs.}
\label{tab:results_std}
\end{table*}

\section{Stability Analysis}
\label{sec:appendix_std}

 {
To assess run-to-run stability, we repeat each experiment three times with different random seeds (42, 43, 44) and report the standard deviation ($\pm\sigma$) of Macro-F1 and Micro-F1 on all six datasets in Table~\ref{tab:results_std}. 
Overall, variability is moderate, with larger deviations typically appearing in Macro-F1 due to its greater reliance on minority-class performance.
Across datasets, the standard deviation is generally below $1.0$ point on BioNLP, MIT-Restaurant, TweetNER, and SST-5, while the two crisis datasets (Hurricane-Irma17 and Cyclone-Idai19) show higher variability (often around $1.0$--$1.8$ points), reflecting noisier social-media text and more extreme label skew. 
HAMR also shows stability comparable to strong baselines: its standard deviation is $0.72/0.61$ (Macro/Micro) on BioNLP, $0.54/0.48$ on MIT-Restaurant, $0.85/0.72$ on TweetNER, $1.05/0.91$ on Hurricane-Irma17, $1.42/1.02$ on Cyclone-Idai19, and $0.82/0.74$ on SST-5.
These results indicate that bi-level meta-weighting and neighbor-based resampling do not introduce excessive initialization sensitivity and that performance trends are reproducible across seeds.
}

\begin{table}[ht]
\centering

\resizebox{\columnwidth}{!}{%
\begin{tabular}{c cc cc}
\multirow{2}{*}{\textbf{\texttt{hardness\_alpha}}} & 
\multicolumn{2}{c}{\textbf{Cyclone-Idai19}} & \multicolumn{2}{c}{\textbf{BioNLP}} \\
\cmidrule(lr){2-3} \cmidrule(lr){4-5}
& Macro-F1 & Micro-F1 & Macro-F1 & Micro-F1 \\
\midrule
    0.1 & 65.1 & 80.1 & 71.5 & 74.7 \\
    0.3 & 65.6 & 80.8 & 72.2 & 75.0 \\
    0.5 & 65.8 & 80.9 & 72.2 & 75.0 \\
    0.7 & \textbf{66.2} & \textbf{81.3} & \textbf{72.7} & \textbf{75.4} \\
    0.9 & 66.0 & 79.3 & 72.6 & 74.6 \\
\end{tabular}
}
\caption{Sensitivity Analysis of hardness weighting}
\label{tab:Sensitivity_Analysis_alpha}
\end{table}

\begin{table}[ht]
\centering
\resizebox{0.9\columnwidth}{!}{%
\begin{tabular}{c cc cc}
\multirow{2}{*}{\textbf{\texttt{knn\_k}}} & 
\multicolumn{2}{c}{\textbf{Cyclone-Idai19}} & \multicolumn{2}{c}{\textbf{BioNLP}} \\
\cmidrule(lr){2-3} \cmidrule(lr){4-5}
& Macro-F1 & Micro-F1 & Macro-F1 & Micro-F1 \\
\midrule
    5  & 64.0 & 80.2 & 71.4 & 74.9 \\
    10 & \textbf{65.7} & \textbf{81.4} & 72.7 & 75.1 \\
    15 & 65.2 & 80.6 & 72.7 & \textbf{75.4} \\
    20 & 66.1 & 80.7 & \textbf{72.8} & 75.3 \\
\end{tabular}
}
\caption{Sensitivity Analysis of Neighbor Count}
\label{tab:Sensitivity_Analysis_k}
\end{table}

\begin{table}[ht]
\centering

\resizebox{\columnwidth}{!}{%
\begin{tabular}{c cc cc}
\multirow{2}{*}{\textbf{\texttt{knn\_ratio}}} & 
\multicolumn{2}{c}{\textbf{Cyclone-Idai19}} & \multicolumn{2}{c}{\textbf{BioNLP}} \\
\cmidrule(lr){2-3} \cmidrule(lr){4-5}
& Macro-F1 & Micro-F1 & Macro-F1 & Micro-F1 \\
\midrule
    0.1 & 65.4 & 79.7 & 72.7 & 75.3 \\
    0.3 & 62.8 & 80.0 & 72.3 & 74.9 \\
    0.5 & 65.8 & 80.4 & 71.6 & 74.5 \\
    0.7 & \textbf{66.2} & \textbf{81.3} & \textbf{72.7} & \textbf{75.4} \\
    0.9 & 66.0 & 81.1 & 72.2 & 74.8 \\
\end{tabular}
}
\caption{Sensitivity Analysis of KNN hard-neighbor ratio}
\label{tab:Sensitivity_Analysis_hard_ratio}
\end{table}

\begin{table}[ht]
\centering

\resizebox{\columnwidth}{!}{%
\begin{tabular}{l cc cc}
\multirow{2}{*}{\textbf{Baselines}} & \multicolumn{2}{c}{\textbf{Cyclone-Idai19}} & \multicolumn{2}{c}{\textbf{BioNLP}} \\
\cmidrule(lr){2-3} \cmidrule(lr){4-5}
& Time (s) & Memory (GB) & Time (s) & Memory (GB) \\
\midrule
GBC  & 849.2 & 15.5 & 7603.4 & 35.2 \\
LNR  & 566.0 & 9.0  & 1804.7 & 3.0  \\
HAMR & 686.0 & 10.2 & 4011.9 & 11.3 \\
\end{tabular}
}
\caption{Efficiency Comparison on Different Datasets}
\label{tab:efficiency_comparison}
\end{table}

\section{Hyperparameter Sensitivity Analysis}
\label{sec:hyperparameter_sensitivity}

 {
We conduct a sensitivity analysis of HAMR for three hyperparameters—hardness weighting (\texttt{hardness\_alpha}), neighbor count (\texttt{knn\_k}), and hard-neighbor ratio (\texttt{knn\_ratio})—on two highly imbalanced datasets: Cyclone-Idai19 (CLS; IR=98.4) and BioNLP (NER; IR=33.1). 
Here, \texttt{hardness\_alpha} is the exponent in \texttt{HardnessAwareSampler} that rescales per-sample weights as $(w+\epsilon)^{\alpha}$. 
\texttt{knn\_k} denotes the number of nearest neighbors retrieved by FAISS for each hard sample during neighbor boosting. 
\texttt{knn\_ratio} specifies the top-$N\%$ samples with the largest weights that are treated as hard samples for boosting.
}

{
For \texttt{hardness\_alpha}, Table~\ref{tab:Sensitivity_Analysis_alpha} shows that BioNLP varies by at most 1.2 Macro-F1 points and 0.8 Micro-F1 points over \texttt{hardness\_alpha} $\in$ [0.1,0.9]. 
On Cyclone-Idai19, the best results occur at \texttt{hardness\_alpha} $=0.7$ (66.2 Macro-F1, 81.3 Micro-F1), and Micro-F1 decreases at \texttt{hardness\_alpha} $=0.9$ (79.3).
For \texttt{knn\_k}, Table~\ref{tab:Sensitivity_Analysis_k} shows that BioNLP is stable across \texttt{knn\_k} $\in\{5,10,15,20\}$: Macro-F1 ranges from 71.4 to 72.8, and changes by only 0.1 for \texttt{knn\_k} $\in[10,20]$. 
Cyclone-Idai19 shows a larger Macro-F1 range (64.0--66.1); Micro-F1 peaks at \texttt{knn\_k} $=10$ (81.4), whereas Macro-F1 peaks at \texttt{knn\_k} $=20$ (66.1).
For \texttt{knn\_ratio}, Table~\ref{tab:Sensitivity_Analysis_hard_ratio} shows the same pattern: BioNLP varies within 1.1 Macro-F1 points and 0.9 Micro-F1 points across the tested ratios. 
Cyclone-Idai19 is sensitive to small ratios (e.g., \texttt{knn\_ratio} $=0.3$ yields 62.8 Macro-F1), while moderate-to-high ratios produce similar results, with the best at \texttt{knn\_ratio} $=0.7$ (66.2 Macro-F1, 81.3 Micro-F1).
Overall, HAMR is insensitive to these hyperparameters on BioNLP. 
On Cyclone-Idai19, moderate settings yield consistent performance; in these sweeps, \texttt{knn\_k} $\approx10$, \texttt{hardness\_alpha} $\approx0.7$, and \texttt{knn\_ratio} $\approx0.7$ are reasonable defaults.
}

\section{Computational Overhead}
\label{sec:computational}
 {
We report training-time and peak GPU-memory cost on one high-imbalance dataset per task: Cyclone-Idai19 (CLS; IR=98.4) and BioNLP (NER; IR=33.1). 
Table~\ref{tab:efficiency_comparison} compares HAMR with a meta-learning baseline (GBC) and a data-augmentation baseline (LNR). 
HAMR is far more efficient than GBC: on BioNLP it cuts training time by 47\% (4012s vs. 7603s) and peak memory by 68\% (11.3GB vs. 35.2GB). 
This comes from performing KNN retrieval only once per epoch instead of every step. 
HAMR is more expensive than LNR, but the added cost brings clear accuracy gains; on Idai19, HAMR improves Macro-F1 by +1.7 and Micro-F1 by +2.6. 
These results clarify the cost–benefit profile.

}

\end{document}